\documentclass[sigconf]{acmart}

\usepackage{booktabs} 
\usepackage{bm}
\usepackage{url}
\usepackage{mathrsfs}
\usepackage{multirow}
\usepackage{balance}
\usepackage{amsthm}
\usepackage{amsmath}
\usepackage{amstext}
\usepackage{amssymb}
\usepackage[ruled,vlined,linesnumbered]{algorithm2e}
\usepackage{subcaption}
\usepackage{overpic}
\usepackage{color}
\usepackage{enumitem}

\AtBeginDocument{%
  \providecommand\BibTeX{{%
    \normalfont B\kern-0.5em{\scshape i\kern-0.25em b}\kern-0.8em\TeX}}}

\copyrightyear{2020}
\acmYear{2020}
\setcopyright{acmcopyright}\acmConference[CIKM '20]{Proceedings of the 29th ACM International Conference on Information and Knowledge Management}{October 19--23, 2020}{Virtual Event, Ireland}
\acmBooktitle{Proceedings of the 29th ACM International Conference on Information and Knowledge Management (CIKM '20), October 19--23, 2020, Virtual Event, Ireland}
\acmPrice{15.00}
\acmDOI{10.1145/3340531.3411892}
\acmISBN{978-1-4503-6859-9/20/10}

\begin{document}
\fancyhead{}
\title{Towards Generalizable Deepfake Detection \\ with Locality-Aware AutoEncoder}
\author{Mengnan Du*, Shiva Pentyala*, Yuening Li, Xia Hu}
\thanks{*These authors contributed equally}
\affiliation{
 \institution{Department of Computer Science and Engineering, Texas A\&M University} 
 \city{} 
 }
\email{{dumengnan, pk123, liyuening, xiahu}@tamu.edu}

\ccsdesc[500]{Computing methodologies~Object recognition} 

\begin{abstract}
With advancements of deep learning techniques, it is now possible to generate super-realistic images and videos, i.e., deepfakes. These deepfakes could reach mass audience and result in adverse impacts on our society. Although lots of efforts have been devoted to detect deepfakes, their performance drops significantly on previously unseen but related manipulations and the detection generalization capability remains a problem. Motivated by the \emph{fine-grained} nature and \emph{spatial locality} characteristics of deepfakes, we propose \emph{Locality-Aware AutoEncoder (LAE)} to bridge the generalization gap. In the training process, we use a pixel-wise mask to regularize local interpretation of LAE to enforce the model to learn intrinsic representation from the forgery region, instead of capturing artifacts in the training set and learning superficial correlations to perform detection. We further propose an active learning framework to select the challenging candidates for labeling, which requires human masks for less than 3\% of the training data, dramatically reducing the annotation efforts to regularize interpretations. Experimental results on three deepfake detection tasks indicate that LAE could focus on the forgery regions to make decisions. The analysis further shows that LAE outperforms the state-of-the-arts by 6.52\%, 12.03\%, and 3.08\% respectively on three deepfake detection tasks in terms of generalization accuracy on previously unseen manipulations. 
\end{abstract}

\keywords{Deepfake Detection; GAN; Generalization; Interpretation}

\maketitle

\section{Introduction}
Recently advanced deep learning and computer vision techniques, e.g., generative adversarial networks (GAN), have enabled generation of super-realistic fake images and videos, known as \emph{deepfakes}. These techniques enable attackers or even lay users of machine learning to manipulate an image/video by swapping its content with alternative contents and synthesize a new image/video. For instance, FaceSwap could generate forged videos about real people performing fictional things, where even humans have difficulty in distinguishing these forgeries from real ones~\cite{deep2019,faceswap2019}. In this paper, we employ the broad definition of deepfakes, and do not limit it to facial manipulations. We also consider general GAN-based inpainting manipulation as deepfakes, since it is also fake contents generated by deep learning techniques. The deepfakes could be further shared on social media for malicious purposes, such as spreading fake news, influencing elections or manipulating stock prices, thus could cause serious negative impact on our society~\cite{cozzolino2018forensictransfer}. To help mitigate the adverse effects, it is essential that we develop methods to detect the manipulated forgeries~\cite{agarwal2019protecting,cfdc2019}.

Current deepfake detection approaches are usually formulated as a binary classification problem, which roughly falls into two categories: hand-crafted feature based methods and deep learning based methods. 
The first category is based on the philosophy that deepfakes are generated by algorithms rather than real camera, thus lots of clues and artifacts could be detected. It relies on hypothesis on artifacts or inconsistencies of a video or image, such as lack of realistic eye blinking~\cite{li2018ictu}, face warping artifacts~\cite{li2018exposing}, and lacking self-consistency~\cite{huh2018fighting}. Handcrafted features are thereafter created to detect forgeries. In contrast, the second category develops deep learning models to automatically extract discriminative features to detect forgeries. Those methods take either the whole or partial image as input and then classify it as fake or not by designing various architectures of convolutional networks~\cite{nguyen2019capsule,afchar2018mesonet}. 

Despite abundant efforts have been devoted to forensics, it remains a challenging problem to develop deepfake detection methods which have high generalization capability on previously unseen forgeries. Firstly, as evidenced by recent work~\cite{cozzolino2018forensictransfer,khodabakhsh2018fake}, although current methods could achieve 99\% accuracy on hold-out test set for most tasks, the performance drops to around 50\% random guess accuracy on previously unseen forgery images/videos. 
Secondly, in our preliminary experiments we have observed that these models fail to focus on the forgery regions to make detection, leveraging heatmaps by local interpretation methods~\cite{selvaraju2017grad,zhou2016learning}. Instead, they have concentrated on non-forgery parts, and learn superficial and easier-to-learn correlations to separate true and fake images. Due to the independent and identically distributed (i.i.d.) training-test split of data, these superficial patterns happen to be predictive on hold-out test set. In contrast, forgeries generated by alternative methods may not contain these superficial correlations. This can possibly explain their high accuracy on hold-out test set and low accuracy on alternative test set. Thirdly, to some extent these methods have solved the training dataset, but it is still too far away from really solving the deepfake detection problem. As new types of deepfake manipulations emerge quickly, unfortunately the forensic methods without sufficient generalization capability cannot be readily applied to real world data~\cite{xuan2019generalization,khodabakhsh2018fake}.

To bridge the generalization gap, we propose to investigate the following two distinct characteristics of the deepfakes. Firstly, deepfake detection is \emph{a fine-grained classification task}. The difference between true and fake images is so subtle, even human eyes are hard to distinguish them. Secondly, deepfakes usually have \emph{spatial locality}, where forgery occupies a certain ratio of the whole image input. For instance, DeepFake videos~\cite{deep2019} use GAN-based technology to replace one's face with the other. This manipulation changes human faces, while leaving the background part unchanged. Considering these two properties, a desirable detection model should be able to concentrate on the forgery region to learn effective representations. Towards this end, the detection model needs to possess local interpretability, which could indicate which region is attended by the model to make decisions. The benefit is that we can control the local interpretation explicitly by imposing extra supervision on instance interpretation in the learning process, in order to enforce the model to focus on the forgery region to learn representations.

In this work, based on aforementioned observations, we develop a Locality-Aware AutoEncoder (LAE) for better generalization of deepfake detection. LAE considers both fine-grained representation learning and enforcing locality in a single framework for image forensics. To guarantee fine-grained representation learning, our work builds upon an autoencoder, which employs reconstruction losses and latent space loss for capturing the distribution for the trained images. 
To suppress the superficial correlations learned by the autoencoder, we augment local interpretability to the antoencoder and use extra pixel-wise forgery ground truth to regularize the local interpretation.
As such, the LAE is enforced to capture discriminative representations from the forgery region. We further employ an active learning framework to reduce the efforts to create pixel-wise forgery masks.
The major contributions of this paper are summarized as follows:
\begin{itemize}[leftmargin=*]
\setlength\itemsep{0em}
\item We propose a deepfake detection method, called LAE, which makes predictions relying on correct evidence in order to boost generalization accuracy.

\item We present an active learning framework to reduce the annotation efforts, where less than 3\% annotations are needed to regularize LAE during training.

\item Experimental results on three deepfake detection tasks validate that LAE could push models to learn intrinsic representation from forgery regions. The proposed LAE achieves state-of-the-art generalization accuracy on previously unseen manipulations. 
\end{itemize}

\section{The Proposed LAE Framework}
\begin{figure*}
  \centering
  \includegraphics[width=0.98\linewidth]{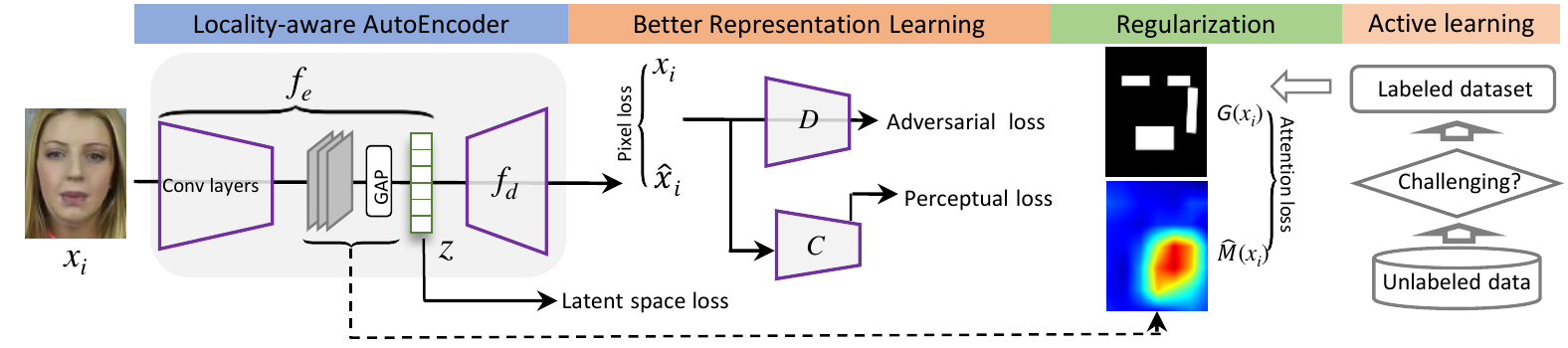}
  \caption{\small{Schematic of LAE training for generalizable deepfake detection. Latent space and reconstruction losses are used to force LAE to learn effective representation. Extra supervision is utilized to regularize local interpretation to boost generalization accuracy. Active learning is exploited to reduce forgery masks annotation efforts.}
  }
  \label{fig:pipeline}
\end{figure*}

In this section, we introduce the proposed framework for generalizable deepfake detection. The pipeline of the proposed framework is illustrated in Fig.~\ref{fig:pipeline}. The key idea is to regularize detection model with external annotations, which could guide the model to learn intrinsic representation (Sec. \ref{sec:autoencoder}) from the right and justified forgery region (Sec. \ref{sec:lae}). Besides, active learning framework is utilized to reduce human annotation efforts for model regularization (Sec. \ref{sec:activelearning}).

\subsection{Problem Statement}  
In this section, we first introduce the basic notations used in this paper. Then we present the generalizable deepfake detection problem that we aim to tackle.

\vspace{3pt}
\noindent\textbf{Notations}: In this paper, we employ a general definition of deepfake, which denotes fake contents generated by advanced deep learning techniques. Representative examples include face swap, facial attributes manipulation and inpainting manipulations. Given a \emph{seen dataset} $\mathcal{D}$ containing both true images $\textbf{X}_T$ and fake images $\textbf{X}_F$ generated by a forgery method. $\mathcal{D}$ is split into training set $\mathcal{D}_{\text{trn}} = \{(x_i, l_i)\}_{i=1}^{N_{\text{trn}}}$, validation set $\mathcal{D}_{\text{val}} = \{(x_i, l_i)\}_{i=1}^{N_{\text{val}}}$ and test set $\mathcal{D}_{\text{tst}} = \{(x_i, l_i)\}_{i=1}^{N_{\text{tst}}}$, where $l_i \in [0,1]$ denotes fake and true class label respectively. A detection model $f(x)$ is learned from the training set $\mathcal{D}_{\text{trn}}$. Besides seen set $\mathcal{D}$, there is also a \emph{unseen dataset} $\mathcal{D}_{\text{unseen}} = \{(x_i, l_i)\}_{i=1}^{N_{\text{unseen}}}$, which is used to test the generalization of model $f(x)$ on unseen manipulations. Fake images in $\mathcal{D}_{\text{unseen}}$ and in $\mathcal{D}$ belong to the same kind of forgery task, while are not generated by the same forgery methods. Take face swap for example: $\mathcal{D}$ contains forgery images created by FaceSwap~\cite{faceswap2019}, while fake images in $\mathcal{D}_{\text{unseen}}$ are created by an alternative forgery method, such as Face2Face~\cite{thies2016face2face}. Besides, the unseen dataset $\mathcal{D}_{\text{unseen}}$ only serves the testing purpose, and none of its images is used for model training, hyperparameters tuning or validation purposes.

\vspace{3pt}
\noindent\textbf{Generalizable Deepfake Detection}: Our objective is to train a model, which could generalize across a large variety of possible manipulations, as long as they are for the same detection task. For instance, for the face manipulation detection task, we expect the model trained on FaceSwap~\cite{faceswap2019} is able to generalize to alternative manipulation methods, such as Face2Face~\cite{thies2016face2face} and other FaceSwap implementations. This is significant in the real-world scenario, since new manipulation methods emerge day by day, and retraining the detector is difficult and even impractical due to the lack of sufficient labeled data from the new manipulation methods.

\subsection{AutoEncoder for Deepfake Detection} \label{sec:autoencoder}
A key characteristic of deepfake detection lies in its fine-grained nature. Thus effective representation is needed for both true and fake images in order to ensure high detection accuracy. As such, we use an autoencoder to learn more distinguishable representations which could separate true and fake images in the latent space.

The autoencoder is denoted using $f$, which consists a sub-network encoder $f_e(\cdot)$ and decoder $f_d(\cdot)$. This encoder maps the input image $x\in \mathbb{R}^{w\times h \times 3}$ to the low-dimensional latent vector space encoding $z \in \mathbb{R}^{d_z}$, where $d_z$ is the dimension of latent vector $z$. Then the decoder remaps latent vector $z$ back to the input space $\hat{x}\in \mathbb{R}^{w\times h \times 3}$.  
\begin{equation}
\begin{aligned}
  z = f_e(x, \theta_e), \quad \hat{x} = f_d(z, \theta_d),
\end{aligned}
\end{equation}
where $\theta_e$ and $\theta_d$ are parameters for the encoder and decoder respectively. To force our model to learn meaningful and intrinsic features, we introduce the latent space loss as well as reconstruction loss.
\begin{equation}
  \mathcal{L}_1(\theta_e, \theta_d, x, l) =  \alpha_1 \mathcal{L}_{\text{rec}} +\alpha_2 \mathcal{L}_{\text{latent}}.
\label{equ:overall}
\end{equation}
These two losses will be elaborated in following sections.

\vspace{3pt}
\noindent\textbf{Latent Space Loss.}\,
We make use of the latent space representation to distinguish the forgery images from the true ones~\cite{cozzolino2018forensictransfer}. The latent space vector is first split into two parts: $T=\{1,...,\frac{d_z}{2} \}$, and $F=\{\frac{d_z+2}{2},..., d_z\}$. The total activation of $x_i$ for the true and fake category respectively are denoted as:
\begin{equation}
  a_{i,T} = \frac{2}{d_z} ||z_{i,c} ||_1, \, c \in T; \, a_{i,F} = \frac{2}{d_z} ||z_{i,c} ||_1, \, c \in F.
\end{equation}
The final latent space loss is defined as follows:
\begin{equation}
\begin{aligned}
  \mathcal{L}_{\text{latent}} = \sum_{i} |a_{i,T} - l_i|+|a_{i,F} - (1-l_i)|,
\end{aligned}
\label{equ:latent}
\end{equation}
where $l_i$ is the ground truth of input image $x_i$. The key idea of this loss is to enforce the activation of the true part: $\{z_{i,c}\}, \, c\in T$ to be maximally activated if the input $x_i$ is a true image, and similarly to increase the fake part $\{z_{i,c}\}, \, c\in F$ activation values for fake image inputs. At testing stage, the deepfake detection is based on the activation value of the latent space partitions. The input image $x_i$ is considered to be true if $a_{i,T} > a_{i,F}$, and vice versa.

\vspace{3pt}
\noindent\textbf{Reconstruction Loss.}\,
To force the fake and true images more distinguishable in the latent space, it is essential to learn effective representations. Specifically, we use reconstruction loss which contains three parts: \emph{pixel-wise loss}, \emph{perceptual loss}, and \emph{adversarial loss}, to learn intrinsic representation for all training samples. The overall reconstruction loss $\mathcal{L}_{\text{rec}}$ is defined as follows:
\begin{equation}
\footnotesize
  \mathcal{L}_{\text{rec}} =  
  \sum_{i}\beta_1 \underbrace{||x_i - \hat{x}_i||_2^2}_{\text{Pixel Loss}} + \beta_2  \underbrace{||C(x_i) - C(\hat{x}_i)||_2^2}_{\text{Perceptual Loss}} + \beta_3  \underbrace{[-\text{log} (D(\hat{x}_i))]}_{\text{Adversarial Loss}}.
\label{equ:rec}
\end{equation}
The pixel-wise loss is measured using mean absolute error (MAE) between original input image $x_i$ and reconstructed image $\hat{x}_i$. For perceptual loss, a pretrained comparator $C(\cdot)$ (e.g., VGGNet~\cite{simonyan2014very}) is used to map input image to feature space: $\mathbb{R}^{w\times h \times 3} \rightarrow \mathbb{R}^{w_1\times h_1 \times d_1}$. Then MAE difference at the feature space is calculated, which represents high-level semantic difference of $x_i$ and $\hat{x}_i$.
In terms of adversarial loss, a discriminator $D(\cdot)$ is introduced aiming to discriminate the generated images $\hat{x}_i$ from real ones ${x}_i$. This subnetwork $D(\cdot)$ is the standard discriminator network introduced in DCGAN~\cite{radford2015unsupervised}, and is trained concurrently with our autoencoder. The autoencoder is trained to trick the discriminator network into classifying the generated images as real. The discriminator $D$ is trained using the following objective:
\begin{equation}
\begin{aligned}
  \mathcal{L}_{D} = -[ \mathbb{E}_{x \sim P_X } [\text{log} D(x)] + \mathbb{E}_{x \sim P_X } [\text{log}(1- D(\hat{x}))] ].
\end{aligned}
\end{equation}
Parameter $\beta_1$,  $\beta_2$, $\beta_3$ are employed to adjust the impact of individual losses. The three losses serve the purpose of ensuring reconstructed image to: 1) be sound in pixel space, 2) be reliable in the high-level feature space, and 3) look realistic respectively. The implicit effect is to force the vector $z$ to learn intrinsic representation which could make it better separate fake and true images. Besides, using three losses instead of using only pixel-wise loss could help stabilize the training in less number of epoches~\cite{dosovitskiy2016generating}.

\subsection{Locality-Aware AutoEncoder (LAE)} \label{sec:lae}
The key idea of LAE is that the model should focus on correct regions and exploit reasonable evidences rather than capture dataset biases to make predictions. Due to the pure data-driven training paradigm, the 
autoencoder developed in last section is not guaranteed to focus on the forgery region to make predictions. 
Instead the autoencoder may capture superficial correlations which happen to be predictive in current dataset. This would lead to decreased generalization accuracy on unseen data generated by alternative forgery methods. In LAE (see Fig.~\ref{fig:pipeline}), we explicitly enforce the model to rely on the forgery region to make detection predictions, by augmenting the model with local interpretability and regularizing the interpretation attention map with extra supervision.

\vspace{3pt}
\noindent\textbf{Augmenting Local Interpretability.} \quad
The goal of local interpretation is to identify the contributions of each pixel in the input image towards
a specific model prediction. The interpretation is illustrated in the format of heatmap (or attention map).
Inspired by the CNN local interpretation method Class Activation Map (CAM)~\cite{zhou2016learning}, we use global average pooling (GAP) layer as ingredient in the encoder, as illustrated in Fig.~\ref{fig:pipeline}. This enables the encoder to output attention map for each input. Let $l$-layer denotes the last convolutional layer of the encoder, and $f_{l,k} (x_i)$ represents the activation matrix at $k$-channel of $l$-layer for input image $x_i$. Let $w_k^c$ corresponds to the weight of $k$-channel to the unit $c$ of latent vector $z$. The CAM attention map for unit $c$ is defined as follows:
\begin{equation}
\small
  M_{c} (x_i) = \sum_{k=1}^{d_l} w_k^c \cdot f_{l,k} (x_i).
\end{equation}
Later we upsample $M_{c} (x_i)$ to the same dimension as the input image $x_i$ using bilinear interpolation. Each entry within $M_{c} (x_i)$ directly indicates the importance of the value at that spatial grid of image $x_i$ leading to the activation $z_c$.
The final attention map $\hat{M} (x_i)$ for an input image $x_i$ is denoted as:
\begin{equation}
\small
  \hat{M} (x_i) =  \sum_{c=1}^{d_F} |z_{i,c}| \cdot M_{c} (x_i) = \sum_{c=1}^{d_F} \sum_{k=1}^{d_l} |z_{i,c}| \cdot w_k^c \cdot f_{l,k} (x_i),
\end{equation}
where $z_{i,c}$ denotes the $c$-th unit of the latent vector $z$ for $x_i$. 
The heatmap is end-to-end differentiable, amenable for training with backpropagation and updating model parameters.

\vspace{3pt}
\noindent\textbf{Regularizing Local Interpretation.} \quad
To enforce the network to focus on the correct forgery region to make detection, a straightforward way is to use instance-level forgery ground truth to regularize the local interpretation. Specifically the regularization is achieved by minimizing the distance between individual interpretation map $\hat{M} (x_i)$ and the extra supervision for all the $N_F$ forgery images. The attention loss is defined as follows:
\begin{equation}
  \mathcal{L}_{\text{attention}}(\theta_e, x, G) =   \sum_{i=1}^{N_F} [ \hat{M} (x_i) - G (x_i)]^2,
\label{equ:attention-all}
\end{equation}
where $G (x_i)$ denotes extra supervision, which is annotated ground truth for forgery. This ground truth is given in the format of pixel-wise binary segmentation mask (see Fig.~\ref{fig:pipeline} for an illustrative example). The attention loss is end-to-end trainable and can be utilized to update the model parameters. Ultimately the trained model could focus on the manipulated regions to make decisions.

\subsection{Active Learning for Regularization} \label{sec:activelearning}
In last section, we introduce regularizing LAE with pixel-wise segmentation masks.
However, generating these masks is extremely time consuming, especially if we plan to label all $N_F$ forgery images. We are interested in employing only a small ratio of data with extra supervision. In this section, we propose an active learning framework to select challenging candidates for annotation.
We will describe below how the active learning works in three steps. 

\vspace{3pt}
\noindent\textbf{Channels Concept Ranking.} \quad
Due to the hierarchical structure of encoder, the last convolutional layer has larger possibility to capture high-level semantic concepts. In our case, we have 512 channels at this layer. A desirable detector could possess some channels which are responsive to specific and semantically meaningful natural part
(e.g., face, mouth, or eye), while other channels may capture concepts related to forgery, (e.g., warping artifacts, or contextual inconsistency). Nevertheless, in practice the detector may rely on some superficial patterns which only exist in the training set to make forgery predictions. Those samples leading to this concept are considered as the most challenging case, since they cause the model to overfit to dataset specific bias and artifacts.

We intend to select out a subset of channels in the last convolutional layer deemed as most influential to the forgery classification decision. The contribution of a channel towards a decision is defined as the channel's average activation scores for an image. Specifically, the contribution of channel $k$ towards image $x_i$ is denoted as: $\{u_{i,k}\}_{k=1}^{d_c}$, where $d_c$ is the number of channels. We learn a linear model based on the $d_c$ concepts to predict the possibility of image $x_i$ to be fake: $p(u_i) = \frac{\text{exp}(w\cdot u_i)}{1+\text{exp}(w\cdot u_i)}$. The loss function is defined as:
\begin{equation}
  \mathcal{L}_{w} =  \sum_{i=1}  [l_i \cdot \text{log} (p(u_i)) + (1-l_i) \cdot \text{log}(1- p(u_i))].
  \label{equ:linear}
\end{equation}
After this training, we select 10 highest components of the optimized linear weight vector $w$ and the corresponding channels are considered as more relevant to the forgery decision.

\setlength{\textfloatsep}{14pt}
\begin{algorithm}[t!]\small
\DontPrintSemicolon
\KwIn{Training data $D = \{(x_i,l_i)\}_{i=1}^N$.} 
 Set hyperparameters $\alpha_1, \alpha_2, \beta_1, \beta_2, \beta_3, \lambda_1, \lambda_2$, learning rate $\eta$, iteration number $max\_iter1, max\_iter2$, epoch index $t=0$;\;
 Initialize autoencoder parameters $\theta_e, \theta_d$;\;
 \While { $t \leq max\_iter1 $}{
$\mathcal{L}_1(\theta_e, \theta_d, x, l) =  \alpha_1 \mathcal{L}_{\text{rec}} +\alpha_2 \mathcal{L}_{\text{latent}}$;\;
$\theta_{e, t+1}, \theta_{d, t+1} = Adam(\mathcal{L}_1(\theta_e, \theta_d, x, l),\eta)$;\;
$t=t+1$;\;
}
Reduce the learning rate: $\eta \gets \frac{\eta}{10}, t \gets 0$;\;
\While { $t \leq max\_iter2 $}{
$\mathcal{L}_{w} =  \sum_{i=1}  [l_i \cdot \text{log} (p(u_i)) + (1-l_i) \cdot \text{log}(1- p(u_i))]$;\;
Select out $N_\text{active}$ images as active candidates;\;
Request labeling pixel-wise masks $\{G(x_i)\}_{i=1}^{N_\text{active}}$;\;
$\mathcal{L}_{\text{attention}}(\theta_e, x, G) =  \sum_{i=1}^{N_\text{active}} [ \hat{M} (x_i) - G (x_i)]^2$;\;
$\mathcal{L}_2(\theta_e, x, l, G) = \lambda_1 \mathcal{L}_{\text{latent}} +\lambda_2 \mathcal{L}_{\text{attention}}$;\;
$\theta_{e, t+1} = Adam(\mathcal{L}_2(\theta_e, x, l, G),\eta)$;\;
$t=t+1$; $\eta \gets \frac{\eta}{10} $ if $t\ \% \ 3 =0$;\;
}
\KwOut{ LAE makes right predictions via right reasons.}
\caption{Locality-Aware AutoEncoder (LAE).}
\label{alg:CREX}
\end{algorithm}

\vspace{2pt}
\noindent\textbf{Active Candidate Selection.} \quad
 After locating the most possible channels corresponding to the forgery prediction, we feed all the $N_F$ fake images to the LAE model. Those who have highest activation value for these top 10 channels are deemed as the challenging case. The key idea for this choice is that these highest activation images are mostly likely to contain easy patterns which can be captured by the model to separate true and fake images, and which are hard to be generalized beyond training and hold-out test set. Thus we would like to request their pixel-wise forgery masks and followed by regularizing them. Based on this criteria, we select out $N_\text{active}$ images as active candidates. The candidates number $N_\text{active}$ is less than 3\% of total images and is empirically shown significant improvement on generalization accuracy. Comparing to the number of total training samples which is larger than 10k, we have dramatically reduced the labelling efforts.

\vspace{2pt}
\noindent\textbf{Local Interpretation Loss.} \quad Equipped with the active image candidates, we request labeling those images for pixel-wise forgery masks $\{G(x_i)\}_{i=1}^{N_\text{active}}$. The attention loss is calculated using the distance between attention map and annotated forgery mask for $N_\text{active}$ candidates, rather than all $N_F$ forgery images as in Eq.(\ref{equ:attention-all}). 
\begin{equation}
  \mathcal{L}_{\text{attention}}(\theta_e, x, G) = \sum_{i=1}^{N_\text{active}} [ \hat{M} (x_i) - G (x_i)]^2. 
\end{equation}
The attention loss in Eq.(11) is further combined with latent space loss in Eq.(4) to update model parameters.
\begin{equation}
  \mathcal{L}_2(\theta_e, x, l, G) = \lambda_1 \mathcal{L}_{\text{latent}} + \lambda_2 \mathcal{L}_{\text{attention}}.
\label{equ:finalattention}
\end{equation}
The overall learning algorithm of LAE is presented in Algorithm 1. We apply a two-stage optimization to derive a generalizable forgery detector. In the first stage, we use $\mathcal{L}_1$ loss in Eq.(\ref{equ:overall}) to learn an effective representation. In the second stage, we need the model to focus on forgery regions to learn better representations. So we exploit the active learning framework to select out challenging candidates to get their pixel-wise forgery masks. Then we reduce the learning rate one-tenth every 3 epoches and fine-tune the parameters of the encoder using the $\mathcal{L}_2$ loss in Eq.(\ref{equ:finalattention}). Note that during training we also add random noise to the input image, in order to prevent model from learning low-level training set specific statistics which are bad at generalization. After training the model and during the testing stage, we use latent space activation in Eq.(3) to distinguish forgery from true ones. The test images are considered to be true if $a_{i,T} > a_{i,F}$, and vice versa.

\section{Experiments}
In this section, we conduct experiments to evaluate performance of LAE and answer the following research questions (\textbf{RQ}s). 

\begin{itemize}[leftmargin=*]
\item \textbf{RQ1} - Does LAE increase the generalization accuracy when processing unseen instances, especially for those produced by alternative methods?
\item \textbf{RQ2} - Does LAE provide better attention maps after augmenting extra supervision in the training process?
\item \textbf{RQ3} - How do different components and hyperparameters affect the performance of LAE?
\item \textbf{RQ4} - Does LAE provide some insights about the detection model as well as training dataset?

\end{itemize}

\subsection{Experimental Setup}
In this section, we introduce the overall experimental setups, including tasks and datasets, baseline methods, networks architectures and implementation details.

\subsubsection{Tasks and Datasets.}\,
 The overall empirical evaluation is performed on three types of deepfake detection tasks. For each task, we use two datasets: \emph{seen} dataset and \emph{unseen} dataset. The seen dataset is split into training, validation and test set, which are used to train the model, tune the hyperparameters and test the model accuracy respectively. In contrast, unseen dataset contains forgery images generated by an alternative method, and is only utilized to assess the true generalization ability of the detection models. Corresponding dataset statistics are given in Tab.~\ref{tab:datastat}. All subsets of the three tasks are balanced, where the ratio of true and fake images are 1:1.
\begin{itemize}[leftmargin=*]
\item \textbf{Face Swap Manipulation } \, 
This task explores human face manipulations, where face of a person is swapped by face of another person in the video. We use videos from Faceforensics++~\cite{rossler2019faceforensics}. The seen dataset is generated using the manipulation method Face2Face~\cite{thies2016face2face}, while the unseen one is obtained via the manipulation method FaceSwap~\cite{faceswap2019}. The videos are compressed using H.264\footnote{\url{https://www.h264encoder.com/}} with quantization parameter set to 23. There are 1000 videos for each of real, seen, unseen datasets. Each dataset is split into 720, 140, 140 for training, validation and testing respectively. Finally, we use 200 frames per video for training and 10 frames per video for validation and testing.

\item \textbf{Facial Attributes Manipulation} \, This manipulation modifies some attributes of the face, such as color of skin or hair, smile, gender, age, etc, based on GAN-based techniques~\cite{cozzolino2018forensictransfer}. Real images from CelebA dataset~\cite{liu2015deep} are modified with two methods: StarGAN~\cite{choi2018stargan} and Glow~\cite{kingma2018glow}, which are chosen to be seen and unseen dataset respectively. All images are 256$\times$256 pixels.

\item \textbf{Inpainting-based Manipulation} \,  Inpainting is also referred as image completion. In this task we consider fake images by two inpainting methods, G$\&$L~\cite{iizuka2017globally} and ContextAtten~\cite{yu2018generative}, consisting seen and unseen dataset respectively. The inpainting is performed to central 128$\times$128 pixels of the original images.  
\end{itemize}

\begin{table}
\footnotesize
\setlength{\tabcolsep}{3pt}
\centering
\caption{\small{Dataset statistics for three tasks: face swap manipulation, facial attribute manipulation, and inpainting-based manipulation. 
}}
\vspace{-1pt}
\scalebox{1}{
\begin{tabular}{l c c c c c c}
\toprule 
& \multicolumn{2}{c}{\textbf{Swap}} & \multicolumn{2}{c}{\textbf{Attribute}} & \multicolumn{2}{c}{\textbf{Inpainting}} \\
\cmidrule(l){2-3} \cmidrule(l){4-5} \cmidrule(l){6-7}
 & Face2face & \footnotesize{FaceSwap} & StarGAN  &Glow  & G$\&$L & ContextAtten\\ 
\midrule 
Train & 288000 & - & 41590  & - & 28000 & -\\ 
Validation & 2800 & - & 11952  & - & 6000 & -\\ 
Test & 2800 & 2800 & 5982  & 5982 & 6000 & 6000\\ 
\bottomrule 
\end{tabular}
}
\label{tab:datastat}
\end{table}

\subsubsection{Baseline Methods.}\,
We compare LAE with six baselines, where all baselines are trained using the same loss optimizations mentioned in the original papers.
\begin{itemize}[leftmargin=*]
\setlength\itemsep{0em}
\item \textbf{SuppressNet}: A generic manipulation detector~\cite{bayar2016deep}. An architecture is specifically designed to adaptively suppress the high-level content of the image. This model uses a constrained convolutional layer followed by two convolutional, two max-pooling and three fully-connected layers. Constrained convolution layer is designed to suppress the high-level contents of the image. We use a learning rate = $10^{-5}$ with batch size of 64.

\item \textbf{ResidualNet}: Residual-based descriptors are used for forgery detection~\cite{cozzolino2017recasting}. This model recasts the hand-crafted Steganalysis features used in the forensic community to a CNN-based network. Basically, these features are extracted as co-occurrences on 4 pixels patterns along horizontal and vertical direction on the residual image, which is obtained after high-pass filtering of the original input image. We set the learning rate and batch size as $10^{-5}$ and 16 respectively.
\item \textbf{StatsNet}: 
This method integrates the computation of statistical feature extraction within a CNN framework~\cite{rahmouni2017distinguishing}. To optimize the feature extraction scheme, this model uses CNN framework with a global pooling layer that computes four statistics (mean, variance, maximum, minimum). We consider the Stats-2L network since this model has the best performance. We use batch size of 64 with learning rates of $10^{-4}$.

\item \textbf{MesoInception}: An inception module based deepfakes detector, where mean square error instead of cross-entropy is used as loss function~\cite{afchar2018mesonet}. This is a CNN-based network specifically designed to detect face manipulations in videos. It uses two inception modules, two convolutional layers with max-pooling, followed by two fully-connected layers at the end. We use batch size of 64 with learning rates of $10^{-3}$.

\item \textbf{XceptionNet}: 
This is a CNN network based on separable convolutions with residual connections~\cite{chollet2017xception}. We use a pretrained network on ImageNet by replacing last fully connected layer with two outputs in order to match our use-case. We use ImageNet weights to initialize all other layers. XceptionNet is trained with batch size of 32 and learning rate of 0.0002. To set up the newly inserted fully connected layer, we fix all weights up to this new layer and pre-train the network for 3 epochs. Finally, we train the network for additional 20 epochs and choose the one with with best accuracy on validation set. 

\item \textbf{ForensicTransfer}: AutoEncoder-based detector is designed to adapt well to novel manipulation methods~\cite{cozzolino2018forensictransfer}. This is an encoder-decoder based architecture with 5 convolution layers in each sub-network. Decoder additionally uses a 2 $\times$ 2 nearest-neighbor up-sampling before each convolution (except the last one) to recover the original size. The latent space (encoder output) has 128 feature maps among which 64 are associated with the real class and 64 with the fake class. We use a learning rate of 0.001 and a batch size of 64. For a fair comparison with others, we use their version that is not fine-tuned on unseen dataset. 

\end{itemize}
For fair comparison, all models are trained on the same training data and tested on the same hold-out test set and unseen test data. We train the models for a maximum of 40 epochs with early stopping (patience=10). Validation loss is used as criteria for early stopping. 

\begin{table}
\centering
\footnotesize
\caption{\small{Network architecture and output shapes.}} 
\vspace{-2pt}
\scalebox{0.96}{
\begin{tabular}{l c c c}
\toprule 
Encoder layer & Output shape & Decoder layer & Output shape  \\ 
\midrule 
Conv2d  & [64, 128,128] & ConvTranspose2d &  [256, 4,4] \\ 
Relu  & [64, 128,128] & BatchNorm2d \& Relu &  [256, 4,4] \\ 
Conv2d  & [128,64,64] & ConvTranspose2d &  [128, 8,8] \\
BatchNorm2d  &[128,64,64] & BatchNorm2d \& Relu & [128, 8,8] \\
Relu  & [128,64,64] & ConvTranspose2d &  [64, 16,16] \\ 
Conv2d  & [256,32,32] & BatchNorm2d \& Relu &  [64, 16,16] \\
BatchNorm2d  & [256,32,32] & ConvTranspose2d &  [32, 32,32] \\
Relu  & [256,32,32] & BatchNorm2d \& Relu &  [32, 32,32] \\ 
Conv2d  & [512,16,16] & ConvTranspose2d &  [16, 64,64] \\
BatchNorm2d  & [512,16,16] & BatchNorm2d \& Relu &  [16, 64,64] \\
Relu  & [512,16,16] & ConvTranspose2d &  [8, 128,128] \\ 
Conv2d  & [512,16,16] & BatchNorm2d \& Relu &  [8, 128,128] \\
Relu  & [512,16,16] & ConvTranspose2d &  [3, 256,256] \\
AvgPool2d  & [512,1,1] & Tanh &  [3, 256,256] \\
Linear  & [128] &  &   \\
\toprule 
\end{tabular}
}
\label{tab:network}
\end{table}

\subsubsection{Network Architectures.}\,
For encoder and decoder, we use a structure similar to U-net~\cite{ronneberger2015u}.
Details about the layers and corresponding output shapes are given in Tab.~\ref{tab:network}. The AvgPool2d corresponds to global average pooling layer, which transforms the [512,16,16] activation layer into 512 dimension vector. After that, we use a Linear layer to turn it into the 128-dimension latent space vector $z$ (see Fig.~\ref{fig:pipeline}). For comparator $C(\cdot)$, we use the 16-layer version VGGNet~\cite{simonyan2014very}, and the activation after 10-th convolutional layer with output shape [512,28,28] is used to calculate the perceptual loss. For discriminator $D(\cdot)$, we use the standard discriminator network introduced in DCGAN~\cite{radford2015unsupervised}.

\subsubsection{Implementation Details of LAE}\, For image pre-processing, we 
augment Gaussian noise to the input image, to prevent model from learning low level statistics which are bad at generalization. Standard deviation is randomly set to a value between 0 and 5 for each batch. Besides, the noise is only added during training time. The Adam optimizer~\cite{kingma2014adam} is utilized to optimize these models with betas of 0.9 and 0.999, epsilon set to $10^{-8}$, and batch size set as 64. For all tasks, the learning rate is fixed as 0.001 for the first stage in Algorithm 1. Later during finetuning, we freeze the parameters of decoder and discriminator, and only finetune encoder network parameters in the second learning stage of Algorithm 1. We reduce the learning rate by one-tenth every 3 epoches during finetuning stage. Number of finetuning epochs depends on number of active fake images. For instance, 4 and 7 epochs work well for 100 and 500 active images respectively. For the first two hyperparameters ($\beta_1,\beta_2$) in Eq.(5), we have tuned values between 0 and 1 with 0.1 as interval and for the third ($\beta_3$), we have tried \{0.0001, 0.001, 0.01, 0.05, 0.1, 0.5, 1.0\}. During finetuning, we have tried values between 0 and 1 with 0.1 as interval for $\lambda_1,\lambda_2$ in Eq.(12).  Ultimately,
 the following ones work well for all three tasks: $\alpha_1$=1.0, $\alpha_2$=1.0,  $\beta_1$=1.0, $\beta_2$=1.0, $\beta_3$=0.01, $\lambda_1$=0.5, $\lambda_1$=1.0. We apply normalization with mean (0.485, 0.456, 0.406) and standard deviation (0.229, 0.224, 0.225). Besides, unseen dataset only serves testing purposes, and none of images is used to train model or tune hyperparameters.

\subsubsection{Linear Model in Active Learning}
For linear model mentioned in Eq.(10), we use flattened output of Encoder's AvgPool2d layer as input features. Thus every image input to linear model would be represented with 512 features. We train this linear model for 5 epochs with SGD optimizer and 0.001 as learning rate.
The linear model accuracy in hold-out test sets is 96.92\%, 100.0\%, and 99.74\% respectively for Face2face, StarGAN and G\&L datasets.

\begin{figure}
  \centering
  \includegraphics[width=0.9\linewidth]{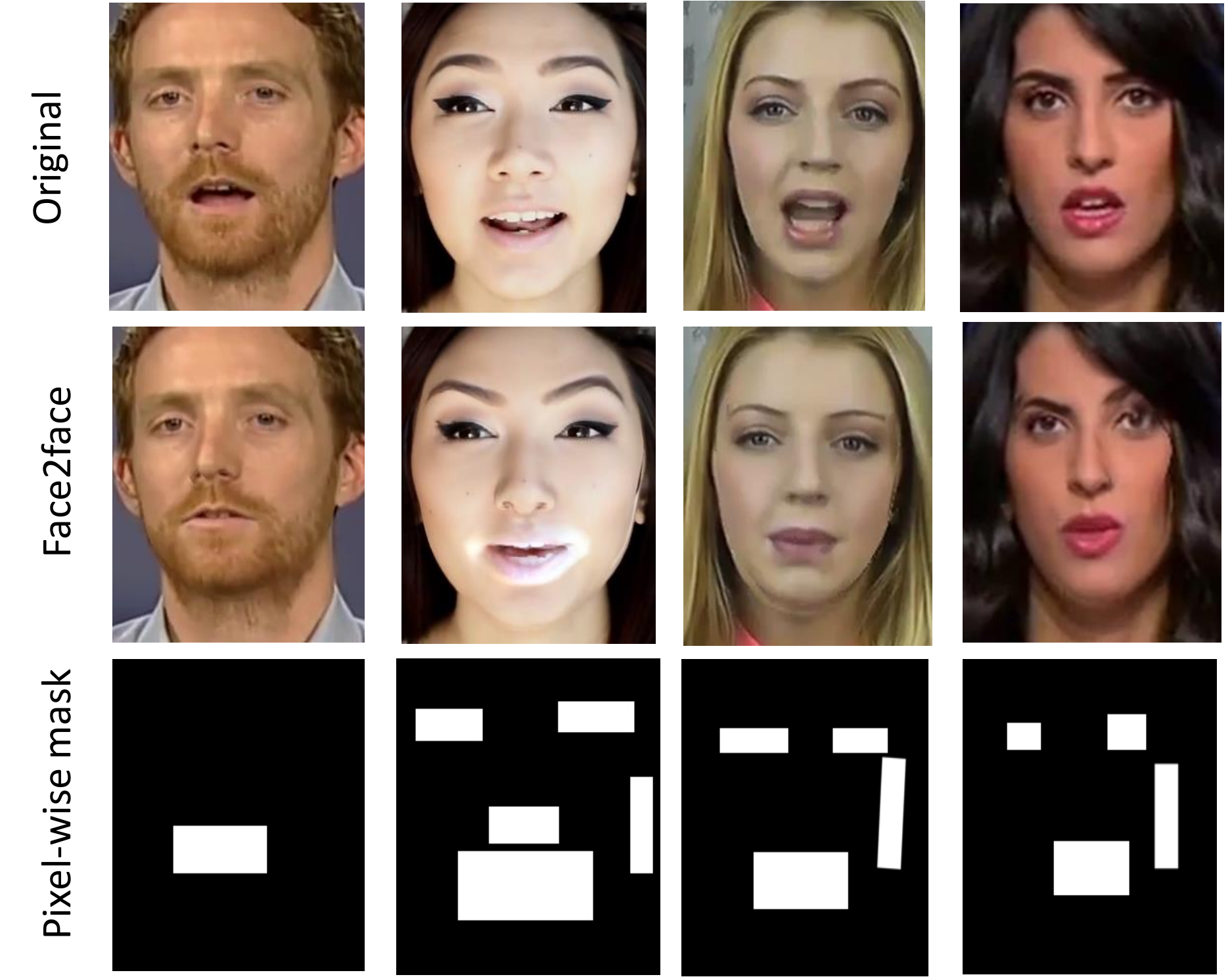}
  \caption{\small{Pixel-wise ground truth masks. The first row displays original face images, the second row shows Face2face manipulations, and the third row represents pixel-wise ground truth masks.}}
  \label{fig:masks}
\end{figure}

\subsubsection{Pixel-wise Masks}
We illustrate some examples of pixel-wise forgery masks in Fig.~\ref{fig:masks}. These masks are for training Face2face dataset of face swap manipulation task, and give the detailed manipulated regions. Specifically, the white color region denotes manipulated parts, while black color regions represent pristine parts.

\begin{table}
\small
\centering
\setlength{\tabcolsep}{3pt}
\caption{\small{Detection accuracy on hold-out test set of \emph{seen dataset} and generalization accuracy on \emph{unseen dataset}. 
}}
\scalebox{0.9}{
\begin{tabular}{|l |c c |c c |c c|}
\toprule 
& \multicolumn{2}{c|}{\textbf{Swap }} & \multicolumn{2}{c|}{\textbf{Attribute }} & \multicolumn{2}{c|}{\textbf{Inpainting}} \\
\cmidrule(l){2-3} \cmidrule(l){4-5} \cmidrule(l){6-7}
\textbf{Models} & Face2face & FaceSwap & StarGAN  &Glow  & G$\&$L & ContextAtten\\ 
\midrule 
SuppressNet      & 93.86 & 50.92 & 99.98  & 49.94 & 99.08 & 49.98\\ 
ResidualNet      & 86.67 & 61.54 & 99.98  & 49.86 & 98.96 & 58.45\\ 
StatsNet         & 92.94 & 57.74 & 99.98  & 50.04 & 96.17 & 50.12\\ 
MesoInception    & 94.38 & 47.32 & 100.0  & 50.01 & 86.90 & 61.34\\ 
XceptionNet      & 98.02 & 49.94 & 100.0  & 49.67 & 99.86 & 50.16\\ 
ForensicTransfer & 93.91 & 52.81 & 100.0  & 50.08 & 99.65 & 50.05\\ 
\midrule
LAE\_100         & 96.84 & 61.09 & 99.91  & 59.05 & 99.04 & 57.64\\
LAE\_400         & 96.82 & 65.24 & 99.75  & 60.08 & 98.95 & 60.67\\
LAE\_800         & 96.80 & \textbf{68.06} & 99.67  & \textbf{62.11} & 98.94 & \textbf{64.42}\\
\bottomrule 
\end{tabular}
}
\label{tab:generalizationEvaluation}

\end{table}

\subsection{Generalization Accuracy Evaluation}
In this section, we evaluate the generalization performance of detection models.
For three tasks, detection accuracy on hold-out test set and data generated by alternative methods (unseen) are given in Tab.~\ref{tab:generalizationEvaluation}. We summarize some key findings. 

\subsubsection{Generalization Gap.}\,
There is a dramatic accuracy gap between seen and unseen dataset. All baseline methods have relatively high accuracy on hold-out test set (most of them are over 90\%), while having random classification (around 50\%) on unseen dataset. On one hand, this indicates strong overfitting of these models to the superficial patterns in the training set. On the other hand, it reveals the limitation of evaluation schemes in existing literature. Usually the detection performance is calculated using the prediction accuracy on the test set. 
Due to the independent and identically distributed (i.i.d.) training-test split of data, especially in the presence of strong priors, detection model can succeed by simply recognize patterns that happen to be predictive on instances over the test set. This is problematic, and thus test set might fail to adequately measure how well detectors perform on new and previously unseen inputs. 
As new types of forgery emerge quickly, it is recommended for detectors to adopt stronger evaluation protocols.

\begin{figure}
  \centering
  \includegraphics[width=0.96\linewidth]{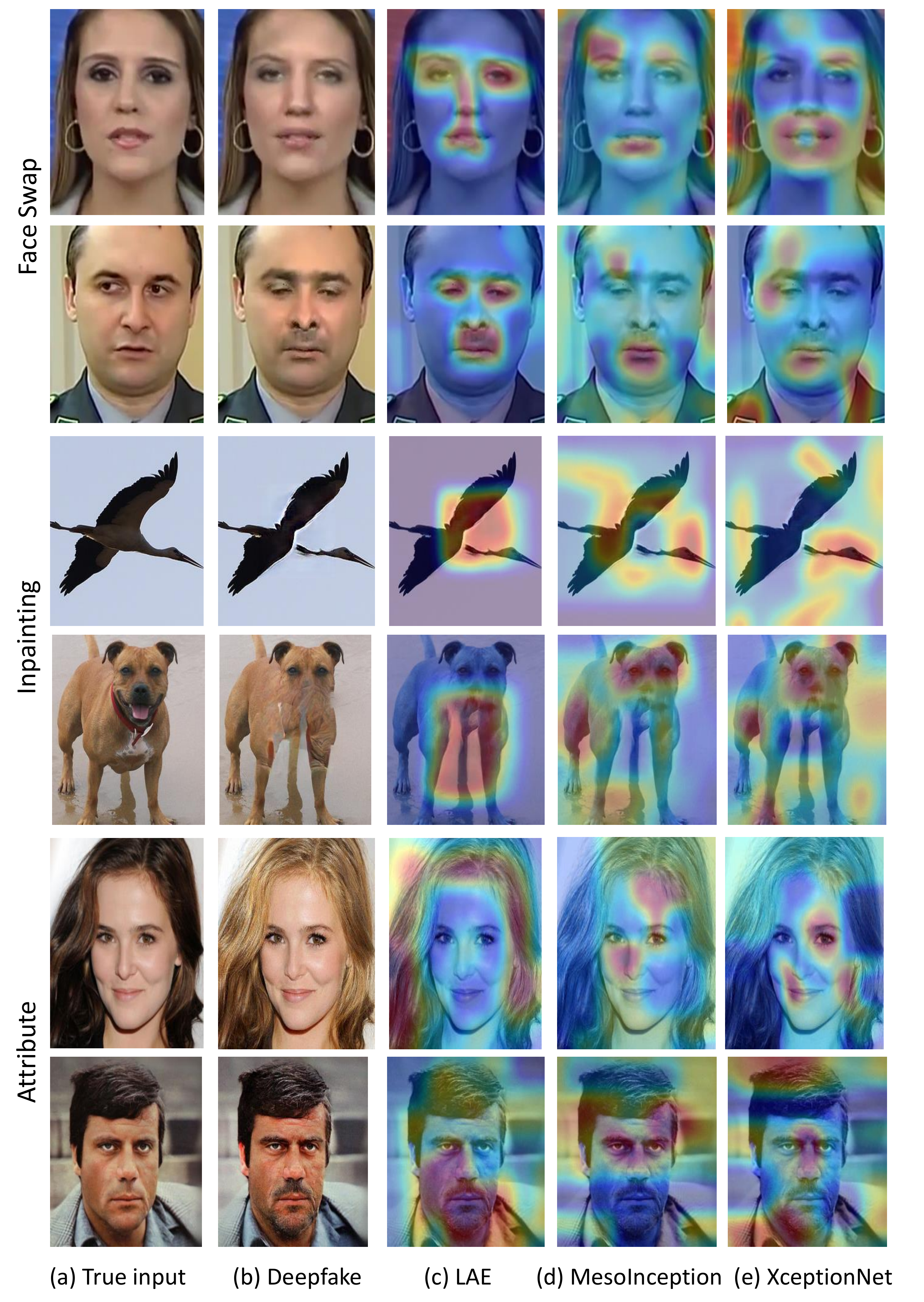}
  \vspace{-3pt}
  \caption{\small{Attention map comparison with baselines. }}
  \label{fig:attention-main}
  \vspace{-3pt}
\end{figure}

\subsubsection{LAE Reduces Generalization Gap.}\,
There are three observations in terms of performance of LAE. Firstly, 
LAE reduces the generalization gap by using a small ratio of extra supervision. LAE\_100, LAE\_400 and LAE\_800 mean the number $N_\text{active}$ is set as 100, 400 and 800 respectively. When using 400 annotations (less than 2\% than total number of training data in Tab.~\ref{tab:datastat}), we achieve state-of-the-art performance on face swap and attribute manipulation tasks. LAE outperforms best baselines by 3.7\% and 10\% respectively on unseen dataset of two tasks. Secondly, using more annotations could bring better generalization enhancement. Compared to 400 annotations, using 800 annotations has boosted the detection accuracy on all three tasks. The generalization accuracy on unseen set has been improved by 6.52\%, 12.03\%, and 3.08\% respectively comparing to best baselines on three deepfake detection tasks. This indicates that LAE has potential to further promote generalization accuracy with more annotations. Thirdly, the increase of generalization accuracy does not sacrifice the accuracy on hold-out test set. Our accuracy on Face2face, StarGAN and G\&L is comparable to baseline methods.

\subsection{Attention Maps Evaluation} 
In this section,  we provide case studies to qualitatively illustrate the effectiveness
of the generated explanation using attention maps for all three deepfake detection tasks.

\subsubsection{Comparison with Baselines.}\,
LAE attention maps are compared with two baselines: MesoInception and XceptionNet (see Fig.~\ref{fig:attention-main}). The heatmaps for both baselines are generated using Grad-CAM~\cite{selvaraju2017grad}.
The visualization  indicates that LAE has truly grasped the intrinsic patterns encoded in the forgery part, instead of picking up superficial and undesirable correlation during the training process. For the first two rows (face swap manipulation), LAE could focus attention on eyes, noses, mouths and beards. In contrast, two baselines mistakenly highlight some background region, e.g., collar and forehead. For the third-fourth rows (inpainting) and fifth-sixth rows (facial attribute), LAE correctly focuses on the inpainted eagle neck, central part of dog and the modified hair, beard regions respectively. By comparison, baselines depends more on non-forgery part, e.g., wings and eyes to make detection.

\subsubsection{Effectiveness of Attention Loss.}
\begin{figure}
  \centering
  \includegraphics[width=0.88\linewidth]{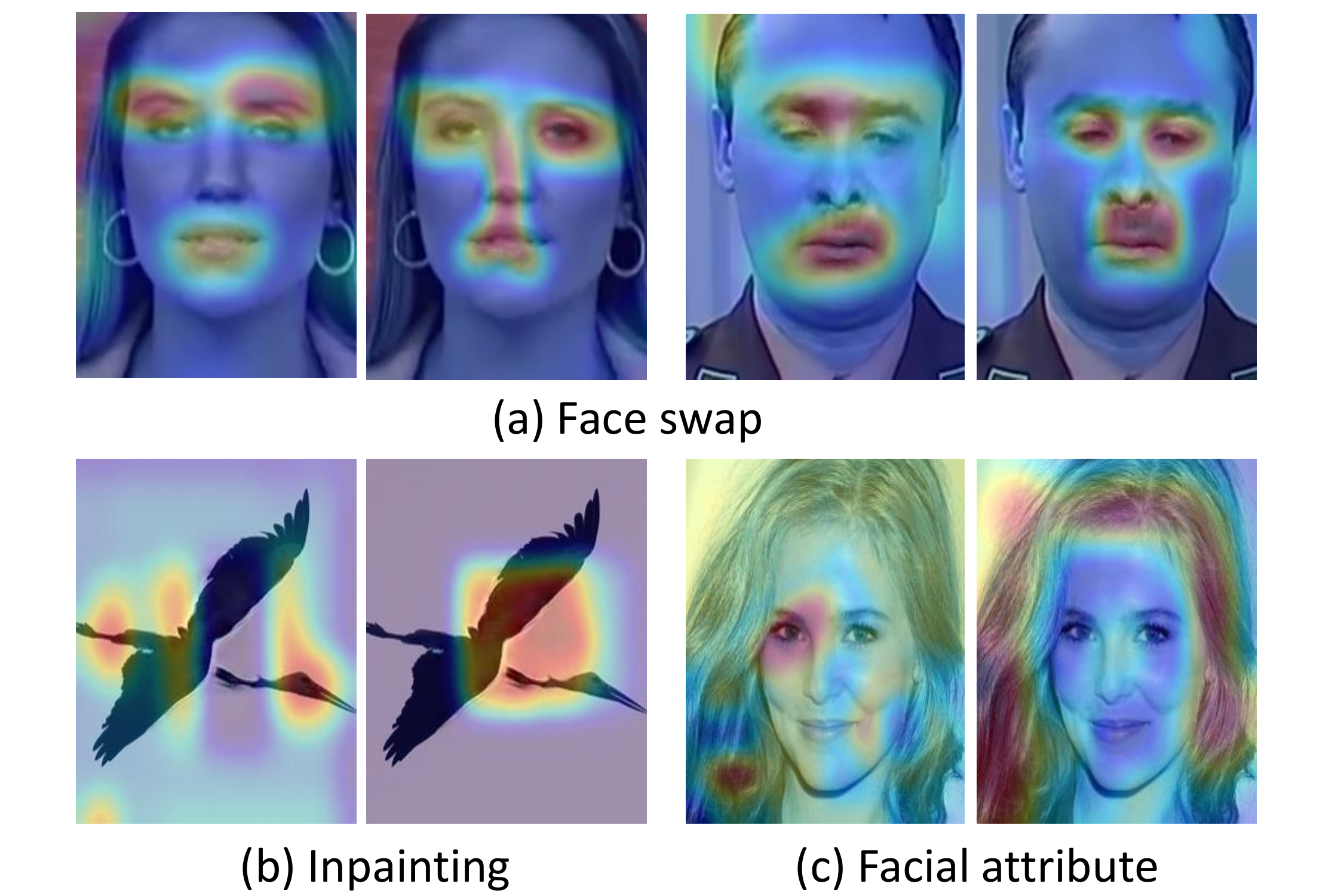}
  \caption{\small{Effectiveness of attention loss. (a) Face swap manipulation, (b) Inpainting, (c) Facial attribute manipulation. For each task, left: before active learning with attention loss, and right: after active learning with attention loss. }}
  \label{fig:attention_loss}
\end{figure}

\,
To qualitatively evaluate the effectiveness of attention loss and active learning, we provide ablation visualizations in Fig.~\ref{fig:attention_loss}. Specifically, we compare LAE\_no\_atten (without using attention loss and active learning) and LAE. For the face swap task, before using attention loss, we can observe that the model does not accurately rely on the forgery region to make decisions. For the female case, LAE\_no\_atten focuses mostly on brow areas, while LAE concentrates correctly on the modified eyes and nose part. For the male case, LAE focuses mostly on the eyes and beard parts, which could mostly distinguish the forgery from real ones. Similarly, for inpainting task and facial attribute manipulation tasks, LAE focuses mostly on the inpainted eagle neck and hair part, comparing to the attention of LAE\_no\_atten on eagle head and female left eye. This validates the effectiveness of attention loss. After finetuning with attention loss with a small ratio of samples provided by active learning, the model eventually concentrates on the forgery part to make predictions.

\subsection{Ablation and Hyperparameters Analysis}
In this section, we utilize models trained on face swap manipulation task to conduct ablation and hyperparameter analysis to study the contribution of different components in LAE.

\begin{table}
\setlength{\tabcolsep}{2.8pt}
\centering
\small
\caption{\small{Ablation analysis of LAE for face swap detection.}} 
\scalebox{0.95}{
\begin{tabular}{l c c c c c}
\toprule 
 & AE\_{rec} & AE\_{latent} & AE\_{latent}\_{pixel} & AE\_{latent}\_{rec} & LAE  \\ 
\midrule 
Face2face &50.39  & 95.82 & 96.57 &  96.92  &  96.80 \\ 
FaceSwap  &49.46  & 50.70 & 50.58 &  50.54  &  68.06 \\ 
\toprule 
\end{tabular}
}
\label{tab:ablation-analysis}
\end{table}

\subsubsection{Ablation Analysis.}\,
We compare LAE with its ablations to identify the contributions of different components. Four ablations include: AE\_{rec}, trained only with reconstruction loss of Eq.(\ref{equ:rec}); AE\_{latent}, using only latent space loss in Eq.(\ref{equ:latent}); AE\_{latent}\_{pixel}, using both latent space loss and pixel loss in Eq.(\ref{equ:rec}); AE\_{latent}\_{rec}, using latent space loss and whole reconstruction loss. Note that no attention loss is used in the ablations. The comparison results are given in Tab.~\ref{tab:ablation-analysis}. There are several key findings. Firstly, latent space loss is the most important part, without which even hold-out test set accuracy could drop to 50.39\%. Secondly, all of pixel-wise, perceptual, and adversarial losses could contribute to performance on hold-out test set. At the same time, no significant increase is observed on the unseen dataset with any combination of these losses. Thirdly, attention loss based on candidates selected via active learning could significantly increase accuracy on unseen dataset (around 17.5\%).

\begin{table}
\setlength{\tabcolsep}{5.5pt}
\centering
\small
\caption{\small{Hyperparameter analysis for $\beta_1,\beta_2,\beta_3$.}} 
\label{tab:beta}
\scalebox{1}{
\begin{tabular}{l c c c c c}
\toprule 
 & $\beta_1$ & $\beta_2$ & $\beta_3$ & Face2face  &  FaceSwap \\ 
\midrule 
Alter pixel & \textbf{1.0}  & 1.0 & 0.01 &  96.92  &  50.54\\ 
 & \textbf{0.5}  & 1.0 & 0.01 &  96.01  &  50.86 \\
 & \textbf{0.1}  & 1.0 & 0.01 &  95.55  &  50.86 \\
\midrule
Alter perceptual  &1.0  & \textbf{1.0} & 0.01 &  96.92  & 50.54\\ 
&1.0  & \textbf{0.5} & 0.01 &  96.74  & 50.82\\
&1.0  & \textbf{0.1} & 0.01 &  95.84  & 50.50\\
\midrule
Alter adversarial  &1.0  &1.0  & \textbf{0.1} &  54.16  & 49.92\\
&1.0  &1.0  & \textbf{0.05} &  58.28  & 50.01\\
&1.0  &1.0  & \textbf{0.01} &  96.92  & 50.54\\
\toprule 
\end{tabular}}
\end{table}
\begin{table}
\setlength{\tabcolsep}{6pt}
\centering
\small
\caption{\small{Hyperparameter analysis for $\lambda_1,\lambda_2$.} }

\scalebox{0.98}{
\begin{tabular}{l c c c c }
\toprule 
 & $\lambda_1$ & $\lambda_2$  & Face2face  &  FaceSwap \\ 
\midrule 
Fix $\lambda_1$=1.0  & 1.0  & 1.0  &  96.85  &  63.14\\ 
 & 1.0  & 0.5  &  96.88  &  55.12 \\
 & 1.0  & 0.1  &  96.90  &  50.61 \\
\midrule
Fix $\lambda_1$=0.5  &0.5  & 1.0  &  96.80  & 68.06\\ 
&0.5  & 0.5  &  96.83  & 62.67\\
&0.5  & 0.1  &  96.88  & 53.05\\
\midrule
Fix $\lambda_1$=0.1  &0.1  &1.0  &  95.97  & 64.92\\
&0.1  &0.5   &  96.42  & 58.01\\
&0.1  &0.1   &  96.85  & 51.06\\
\toprule 
\end{tabular}
}

\label{tab:lambda}
\end{table}

\subsubsection{Hyperparameters Analysis.}\,
We evaluate the effect of different hyperparameters towards model performance by altering the values of $\beta_1,\beta_2,\beta_3$ in Eq.(\ref{equ:rec}) and $\lambda_1,\lambda_2$ in Eq.(\ref{equ:finalattention}). Corresponding results are reported in Tab.~\ref{tab:beta} (without attention loss and active learning) 
and Tab.~\ref{tab:lambda} (with attention loss and active learning) 
respectively. Firstly, the results in Tab.~\ref{tab:beta} indicate that increase of weights for pixel loss and perceptual loss could enhance model performance on test set. In contrast, a small weight for adversarial loss is beneficial for accuracy improvement. Secondly, as shown in Tab.~\ref{tab:lambda}, fixing $\lambda_1$ and reducing $\lambda_2$ from 1.0 to 0.5 then to 0.1 have significantly decreased the accuracy on unseen dataset. This accuracy drop confirms the significance of attention loss in improving generalization accuracy.

\subsubsection{Random vs. Active Learning.}
\begin{figure}
  \centering
  \includegraphics[width=0.95\linewidth]{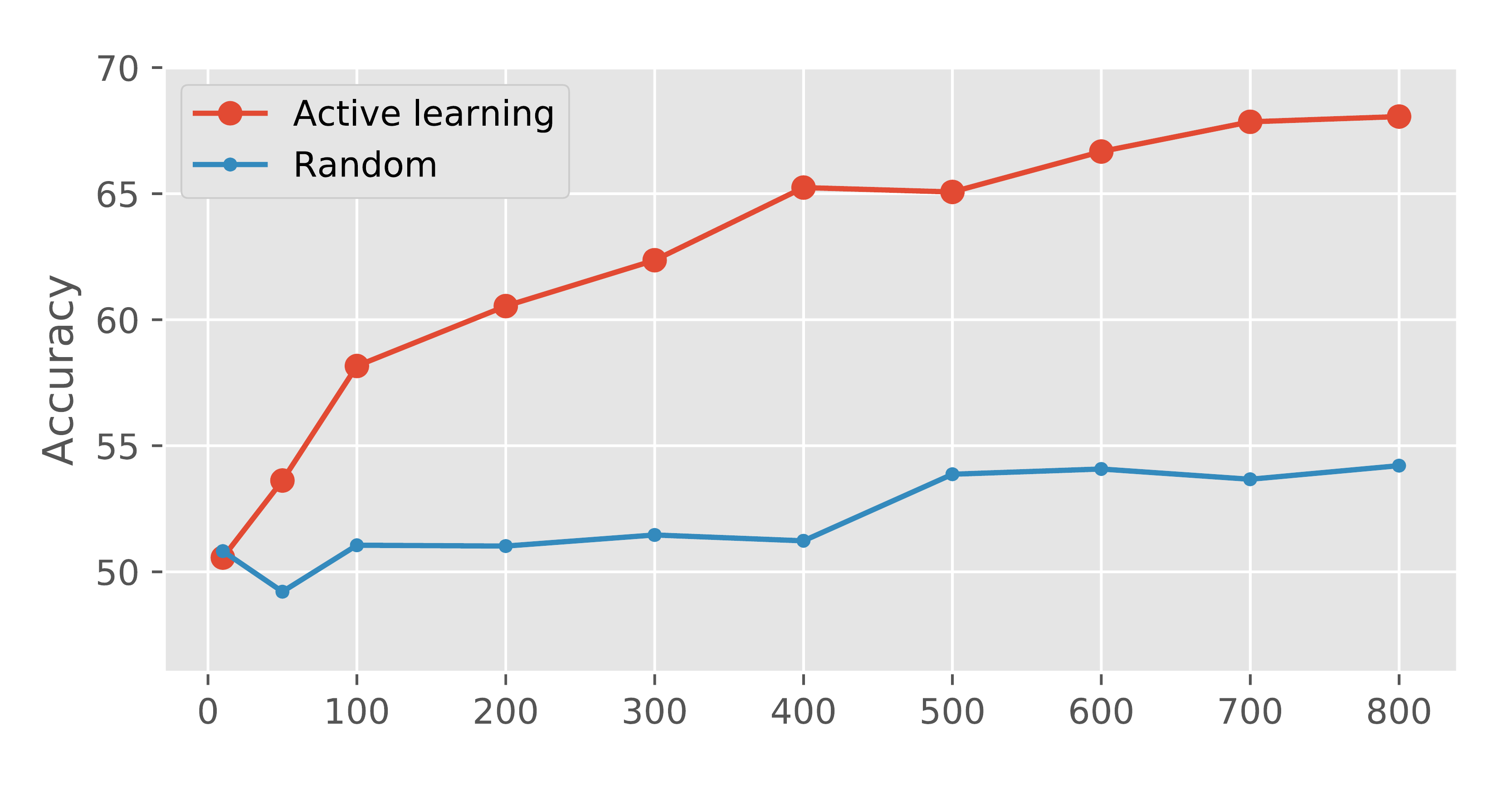}
  \vspace{-4pt}
  \caption{\small{Random and active learning selection comparison. The x axis denotes annotation number which has pixel-wise masks. }}
  \label{fig:random_active}
\end{figure}

\,
We evaluate active learning based challenging candidate selection, by comparing it with random selection. The generalization comparison on unseen dataset (FaceSwap) is illustrated in Fig.~\ref{fig:random_active}. There is a dramatic gap between random selection and active learning. For instance, active learning could increase unseen dataset accuracy by 7.11\% when the annotation number is 100 ($<0.2\%$ of training data). From number 100 to 800, on average active learning could increase accuracy by 11.67\% comparing to random selection. This indicates that active learning is effective in terms of selecting challenging candidates.

\subsubsection{Forgery Ground Truth Number Analysis.}\,
We study the effect of attention regularization by altering the number of challenging candidates($N_\text{active}$) selected by active learning  (see Fig.~\ref{fig:random_active}). There are two main observations. First, increasing the number of annotations typically improves model generalization. Especially using 800 samples has increased the accuracy by 17.5\% than using zero samples, indicating the benefit of extra supervision. Second, using forgery masks for less than $0.2\%$ of training data (100 annotations) has increased accuracy by 7.6\%. Considering the annotation effort of pixel-wise masks, this advantage of requiring small ratio of forgery mask annotations is significant.

\subsection{Debugging Model and Dataset}
In this section, we provide further analysis for the detection task, to get some insights about how to further improve generalization. Specifically, we use interpretability as a debugging tool to analyze our LAE as well as the datasets. The statistical analysis of the attention maps has provided some clues about the weakness of our model and thus possible solutions for improvement.

\subsubsection{Superficial Patterns Captured by Model.}\, 
Since we only use a small ratio of annotated samples (less than 3\%) to regularize LAE, there is still generalization gap on unseen dataset (Tab.~\ref{tab:generalizationEvaluation}). Deepfake detection is a fine-grained classification task, and thus is prone to capture superficial patterns existing in training set. The analysis of attention maps for failure cases of LAE provides some insights about the superficial patterns. Firstly, the model focuses on some semantically meaningful part of object of interest. However, these patterns usually fall in spuriously correlated background region, rather than the true forgery region. Secondly, the model has captured some low-level patterns, such as textures. Both categories happen to be predictive in hold-out test set, while performing poorly on unseen sets. It indicates more inductive bias is needed for architectures and stronger regularization is needed for model training.

\subsubsection{Seen and Unseen Difference.}\,
Through attention map visualizations, we also observe the distribution difference of seen and unseen dataset. For example in face swap detection task, Face2face mainly changes lips and eye brows, while FaceSwap changes mostly nose and eyes. This validates the distribution difference between seen and unseen dataset and brings challenges to generalization accuracy. The accuracy increase bound of LAE depends on the distribution difference between seen dataset and unseen dataset. Towards this end, using a small number of unseen dataset data to finetune model could possibly further reduce the generalization gap, and this direction would be explored in our future research.

\section{Related Work}
In this section, we briefly review three lines of research which are most relevant to ours.

\vspace{3pt}
\noindent\textbf{Traditional Manipulated Forgery Detection } \quad
Techniques for traditional manipulated forgery detection have existed for a long history~\cite{farid2009image}. These forgeries usually are created using traditional image processing techniques. Two typical examples are \emph{image splicing} and \emph{copy-move}, which is the process of cutting one part of a source image, such as the face regions, and inserting it in the target image. 
State-of-the-art methods for traditional forgery detection usually rely on artifacts introduced by forgery generation pipeline to differentiate between forgeries and reals ones, and at the same time could localize forgery regions~\cite{huh2018fighting,wu2019mantra}.

\vspace{3pt}
\noindent\textbf{Generalization of Deepfake Detection } \quad
Deepfake is defined as forgeries generated by advanced deep learning and computer vision techniques~\cite{tolosana2020deepfakes}. Note that traditional manipulated forgeries typically have some artifacts, which enable detection algorithms easily discriminate forgeries with real ones. In contrast, deepfakes usually employ advanced techniques, such as GAN, perceptual loss and adversarial loss to smooth out artifacts~\cite{nguyen2019deep,li2019face}. This generates forgeries which leave almost no clues for detection. Detecting deepfakes is more challenging than traditional forgeries, and is thus the focus of this paper. There are generally two categories of methods for deepfake detection: hand-crafted features based methods, and deep learning based methods~\cite{matern2019exploiting}. The first one relies on high-level visual artifacts of the forgeries, such as lack of realistic eye blinking~\cite{li2018ictu}, and face warping artifacts~\cite{li2018exposing}. However, these models would fail when these hypothesis do not hold. The second category typically formulates deepfake detection as supervised binary classification problem: designing diverse CNN architectures and using end-to-end training for detection. However, these methods tend to perform poorly on new unseen manipulations, where the performance reduces to around 50\% random guess accuracy for previous unseen data~\cite{cozzolino2018forensictransfer}. 

The most similar work to ours is ForensicTransfter method~\cite{cozzolino2018forensictransfer}. Both methods focus on enhancing the generalization of detection methods on previously unseen forgeries. However, their method requires samples from unseen forgeries to finetune detection model, while our method does not require any data from unseen forgeries.

\vspace{3pt}
\noindent\textbf{CNN Local Interpretation} \quad
Local interpretation aims to identify the contributions of each feature in the input towards
a specific DNN prediction~\cite{du2018techniques}. The final interpretation is illustrated in the format of feature importance visualization~\cite{du2018towards,du2019on}. Most current research focus on providing interpretation for opaque DNN models, and these work have exposed some limitations of DNN models. For instance, some models are heavily driven by superficial patterns in data, rather than capturing useful representations. This motivates us to make use of DNN interpretation as an ingredient to promote model's generalization. We use a decomposition based explanation method CAM ~\cite{zhou2016learning}, since it is end-to-end differentiable, amenable for training with backpropagation and updating CNN parameters.

\section{Conclusions and Future Work}
We propose a new deepfake detection method, called Locality-Aware AutoEncoder (LAE), to boost the generalization accuracy by making predictions relying on correct forgery evidence. A key characteristic of LAE is the augmented local interpretability, which could be regularized using extra pixel-wise forgery masks, in order to learn intrinsic and meaningful forgery representations. We also present an active learning framework to reduce the efforts to get forgery masks (less than 3\% of training data). Extensive experiments conducted on three deepfake detection tasks show that our resulting models have a higher probability to look at forgery region rather than unwanted bias and artifacts to make predictions. Empirical analysis further demonstrates that LAE outperforms best baselines by 6.52\%, 12.03\%, and 3.08\% respectively on three deepfake detection tasks on previously unseen manipulated forgeries.

Due to the inherent difficulty of the detection problem, we still could observe generalization gap between test set and unseen dataset that is generated by alternative methods. Although they are related and belong to the same task, there remains slight distribution differences between them. To further reduce this generalization gap, we would explore more comprehensive framework in future research, by combining transfer learning and other techniques. Besides, currently we focus on deepfake image detection, and would explore deepfake video and audio detection in future research.

\bibliographystyle{ACM-Reference-Format}
\bibliography{detection}


\begin{thebibliography}{39}


\ifx \showCODEN    \undefined \def \showCODEN     #1{\unskip}     \fi
\ifx \showDOI      \undefined \def \showDOI       #1{#1}\fi
\ifx \showISBNx    \undefined \def \showISBNx     #1{\unskip}     \fi
\ifx \showISBNxiii \undefined \def \showISBNxiii  #1{\unskip}     \fi
\ifx \showISSN     \undefined \def \showISSN      #1{\unskip}     \fi
\ifx \showLCCN     \undefined \def \showLCCN      #1{\unskip}     \fi
\ifx \shownote     \undefined \def \shownote      #1{#1}          \fi
\ifx \showarticletitle \undefined \def \showarticletitle #1{#1}   \fi
\ifx \showURL      \undefined \def \showURL       {\relax}        \fi
\providecommand\bibfield[2]{#2}
\providecommand\bibinfo[2]{#2}
\providecommand\natexlab[1]{#1}
\providecommand\showeprint[2][]{arXiv:#2}

\bibitem[\protect\citeauthoryear{Afchar, Nozick, Yamagishi, and Echizen}{Afchar
  et~al\mbox{.}}{2018}]%
        {afchar2018mesonet}
\bibfield{author}{\bibinfo{person}{Darius Afchar}, \bibinfo{person}{Vincent
  Nozick}, \bibinfo{person}{Junichi Yamagishi}, {and} \bibinfo{person}{Isao
  Echizen}.} \bibinfo{year}{2018}\natexlab{}.
\newblock \showarticletitle{Mesonet: a compact facial video forgery detection
  network}. In \bibinfo{booktitle}{\emph{2018 IEEE International Workshop on
  Information Forensics and Security (WIFS)}}.
\newblock


\bibitem[\protect\citeauthoryear{Agarwal, Farid, Gu, He, Nagano, and
  Li}{Agarwal et~al\mbox{.}}{2019}]%
        {agarwal2019protecting}
\bibfield{author}{\bibinfo{person}{Shruti Agarwal}, \bibinfo{person}{Hany
  Farid}, \bibinfo{person}{Yuming Gu}, \bibinfo{person}{Mingming He},
  \bibinfo{person}{Koki Nagano}, {and} \bibinfo{person}{Hao Li}.}
  \bibinfo{year}{2019}\natexlab{}.
\newblock \showarticletitle{Protecting World Leaders Against Deep Fakes}. In
  \bibinfo{booktitle}{\emph{The Conference on Computer Vision and Pattern
  Recognition (CVPR) Workshop}}.
\newblock


\bibitem[\protect\citeauthoryear{AWS, Facebook, Microsoft, and academics}{AWS
  et~al\mbox{.}}{2019}]%
        {cfdc2019}
\bibfield{author}{\bibinfo{person}{AWS}, \bibinfo{person}{Facebook},
  \bibinfo{person}{Microsoft}, {and} \bibinfo{person}{academics}.}
  \bibinfo{year}{2019}\natexlab{}.
\newblock \showarticletitle{Deepfake Detection Challenge (DFDC)}.
\newblock


\bibitem[\protect\citeauthoryear{Bayar and Stamm}{Bayar and Stamm}{2016}]%
        {bayar2016deep}
\bibfield{author}{\bibinfo{person}{Belhassen Bayar} {and}
  \bibinfo{person}{Matthew~C Stamm}.} \bibinfo{year}{2016}\natexlab{}.
\newblock \showarticletitle{A deep learning approach to universal image
  manipulation detection using a new convolutional layer}. In
  \bibinfo{booktitle}{\emph{ACM Workshop on Information Hiding and Multimedia
  Security}}.
\newblock


\bibitem[\protect\citeauthoryear{Choi, Choi, Kim, Ha, Kim, and Choo}{Choi
  et~al\mbox{.}}{2018}]%
        {choi2018stargan}
\bibfield{author}{\bibinfo{person}{Yunjey Choi}, \bibinfo{person}{Minje Choi},
  \bibinfo{person}{Munyoung Kim}, \bibinfo{person}{Jung-Woo Ha},
  \bibinfo{person}{Sunghun Kim}, {and} \bibinfo{person}{Jaegul Choo}.}
  \bibinfo{year}{2018}\natexlab{}.
\newblock \showarticletitle{Stargan: Unified generative adversarial networks
  for multi-domain image-to-image translation}. In
  \bibinfo{booktitle}{\emph{The Conference on Computer Vision and Pattern
  Recognition (CVPR)}}.
\newblock


\bibitem[\protect\citeauthoryear{Chollet}{Chollet}{2017}]%
        {chollet2017xception}
\bibfield{author}{\bibinfo{person}{Fran{\c{c}}ois Chollet}.}
  \bibinfo{year}{2017}\natexlab{}.
\newblock \showarticletitle{Xception: Deep learning with depthwise separable
  convolutions}. In \bibinfo{booktitle}{\emph{The Conference on Computer Vision
  and Pattern Recognition (CVPR)}}.
\newblock


\bibitem[\protect\citeauthoryear{Cozzolino, Poggi, and Verdoliva}{Cozzolino
  et~al\mbox{.}}{2017}]%
        {cozzolino2017recasting}
\bibfield{author}{\bibinfo{person}{Davide Cozzolino}, \bibinfo{person}{Giovanni
  Poggi}, {and} \bibinfo{person}{Luisa Verdoliva}.}
  \bibinfo{year}{2017}\natexlab{}.
\newblock \showarticletitle{Recasting residual-based local descriptors as
  convolutional neural networks: an application to image forgery detection}. In
  \bibinfo{booktitle}{\emph{ACM Workshop on Information Hiding and Multimedia
  Security}}.
\newblock


\bibitem[\protect\citeauthoryear{Cozzolino, Thies, R{\"o}ssler, Riess,
  Nie{\ss}ner, and Verdoliva}{Cozzolino et~al\mbox{.}}{2018}]%
        {cozzolino2018forensictransfer}
\bibfield{author}{\bibinfo{person}{Davide Cozzolino}, \bibinfo{person}{Justus
  Thies}, \bibinfo{person}{Andreas R{\"o}ssler}, \bibinfo{person}{Christian
  Riess}, \bibinfo{person}{Matthias Nie{\ss}ner}, {and} \bibinfo{person}{Luisa
  Verdoliva}.} \bibinfo{year}{2018}\natexlab{}.
\newblock \showarticletitle{ForensicTransfer: Weakly-supervised Domain
  Adaptation for Forgery Detection}.
\newblock \bibinfo{journal}{\emph{arXiv preprint arXiv:1812.02510}}
  (\bibinfo{year}{2018}).
\newblock


\bibitem[\protect\citeauthoryear{DeepFake}{DeepFake}{2019}]%
        {deep2019}
\bibfield{author}{\bibinfo{person}{DeepFake}.} \bibinfo{year}{2019}\natexlab{}.
\newblock \showarticletitle{https://github.com/iperov/DeepFaceLab}.
\newblock


\bibitem[\protect\citeauthoryear{Dosovitskiy and Brox}{Dosovitskiy and
  Brox}{2016}]%
        {dosovitskiy2016generating}
\bibfield{author}{\bibinfo{person}{Alexey Dosovitskiy} {and}
  \bibinfo{person}{Thomas Brox}.} \bibinfo{year}{2016}\natexlab{}.
\newblock \showarticletitle{Generating images with perceptual similarity
  metrics based on deep networks}. In \bibinfo{booktitle}{\emph{Thirtieth
  Conference on Neural Information Processing Systems (NIPS)}}.
\newblock


\bibitem[\protect\citeauthoryear{Du, Liu, and Hu}{Du et~al\mbox{.}}{2020}]%
        {du2018techniques}
\bibfield{author}{\bibinfo{person}{Mengnan Du}, \bibinfo{person}{Ninghao Liu},
  {and} \bibinfo{person}{Xia Hu}.} \bibinfo{year}{2020}\natexlab{}.
\newblock \showarticletitle{Techniques for interpretable machine learning}.
\newblock \bibinfo{journal}{\emph{Communications of the ACM (CACM)}}
  (\bibinfo{year}{2020}).
\newblock


\bibitem[\protect\citeauthoryear{Du, Liu, Song, and Hu}{Du
  et~al\mbox{.}}{2018}]%
        {du2018towards}
\bibfield{author}{\bibinfo{person}{Mengnan Du}, \bibinfo{person}{Ninghao Liu},
  \bibinfo{person}{Qingquan Song}, {and} \bibinfo{person}{Xia Hu}.}
  \bibinfo{year}{2018}\natexlab{}.
\newblock \showarticletitle{Towards Explanation of DNN-based Prediction with
  Guided Feature Inversion}.
\newblock \bibinfo{journal}{\emph{ACM SIGKDD International Conference on
  Knowledge Discovery and Data Mining (KDD)}} (\bibinfo{year}{2018}).
\newblock


\bibitem[\protect\citeauthoryear{Du, Liu, Yang, and Hu}{Du
  et~al\mbox{.}}{2019}]%
        {du2019on}
\bibfield{author}{\bibinfo{person}{Mengnan Du}, \bibinfo{person}{Ninghao Liu},
  \bibinfo{person}{Fan Yang}, {and} \bibinfo{person}{Xia Hu}.}
  \bibinfo{year}{2019}\natexlab{}.
\newblock \showarticletitle{On Attribution of Recurrent Neural Network
  Predictions via Additive Decomposition}.
\newblock \bibinfo{journal}{\emph{The Web Conference (WWW)}}
  (\bibinfo{year}{2019}).
\newblock


\bibitem[\protect\citeauthoryear{Faceswap}{Faceswap}{2019}]%
        {faceswap2019}
\bibfield{author}{\bibinfo{person}{Faceswap}.} \bibinfo{year}{2019}\natexlab{}.
\newblock \showarticletitle{https://github.com/shaoanlu/faceswap-GAN}.
\newblock


\bibitem[\protect\citeauthoryear{Farid}{Farid}{2009}]%
        {farid2009image}
\bibfield{author}{\bibinfo{person}{Hany Farid}.}
  \bibinfo{year}{2009}\natexlab{}.
\newblock \showarticletitle{Image forgery detection}.
\newblock \bibinfo{journal}{\emph{IEEE Signal processing magazine}}
  (\bibinfo{year}{2009}).
\newblock


\bibitem[\protect\citeauthoryear{Huh, Liu, Owens, and Efros}{Huh
  et~al\mbox{.}}{2018}]%
        {huh2018fighting}
\bibfield{author}{\bibinfo{person}{Minyoung Huh}, \bibinfo{person}{Andrew Liu},
  \bibinfo{person}{Andrew Owens}, {and} \bibinfo{person}{Alexei~A Efros}.}
  \bibinfo{year}{2018}\natexlab{}.
\newblock \showarticletitle{Fighting fake news: Image splice detection via
  learned self-consistency}. In \bibinfo{booktitle}{\emph{European Conference
  on Computer Vision (ECCV)}}.
\newblock


\bibitem[\protect\citeauthoryear{Iizuka, Simo-Serra, and Ishikawa}{Iizuka
  et~al\mbox{.}}{2017}]%
        {iizuka2017globally}
\bibfield{author}{\bibinfo{person}{Satoshi Iizuka}, \bibinfo{person}{Edgar
  Simo-Serra}, {and} \bibinfo{person}{Hiroshi Ishikawa}.}
  \bibinfo{year}{2017}\natexlab{}.
\newblock \showarticletitle{Globally and locally consistent image completion}.
\newblock \bibinfo{journal}{\emph{ACM Transactions on Graphics (ToG)}}
  (\bibinfo{year}{2017}).
\newblock


\bibitem[\protect\citeauthoryear{Khodabakhsh, Ramachandra, Raja, Wasnik, and
  Busch}{Khodabakhsh et~al\mbox{.}}{2018}]%
        {khodabakhsh2018fake}
\bibfield{author}{\bibinfo{person}{Ali Khodabakhsh},
  \bibinfo{person}{Raghavendra Ramachandra}, \bibinfo{person}{Kiran Raja},
  \bibinfo{person}{Pankaj Wasnik}, {and} \bibinfo{person}{Christoph Busch}.}
  \bibinfo{year}{2018}\natexlab{}.
\newblock \showarticletitle{Fake Face Detection Methods: Can They Be
  Generalized?}. In \bibinfo{booktitle}{\emph{2018 International Conference of
  the Biometrics Special Interest Group (BIOSIG)}}.
\newblock


\bibitem[\protect\citeauthoryear{Kingma and Ba}{Kingma and Ba}{2014}]%
        {kingma2014adam}
\bibfield{author}{\bibinfo{person}{Diederik~P Kingma} {and}
  \bibinfo{person}{Jimmy Ba}.} \bibinfo{year}{2014}\natexlab{}.
\newblock \showarticletitle{Adam: A method for stochastic optimization}.
\newblock \bibinfo{journal}{\emph{arXiv preprint arXiv:1412.6980}}
  (\bibinfo{year}{2014}).
\newblock


\bibitem[\protect\citeauthoryear{Kingma and Dhariwal}{Kingma and
  Dhariwal}{2018}]%
        {kingma2018glow}
\bibfield{author}{\bibinfo{person}{Durk~P Kingma} {and}
  \bibinfo{person}{Prafulla Dhariwal}.} \bibinfo{year}{2018}\natexlab{}.
\newblock \showarticletitle{Glow: Generative flow with invertible 1x1
  convolutions}. In \bibinfo{booktitle}{\emph{Thirty-second Conference on
  Neural Information Processing Systems (NeurIPS)}}.
\newblock


\bibitem[\protect\citeauthoryear{Li, Bao, Zhang, Yang, Chen, Wen, and Guo}{Li
  et~al\mbox{.}}{2020}]%
        {li2019face}
\bibfield{author}{\bibinfo{person}{Lingzhi Li}, \bibinfo{person}{Jianmin Bao},
  \bibinfo{person}{Ting Zhang}, \bibinfo{person}{Hao Yang},
  \bibinfo{person}{Dong Chen}, \bibinfo{person}{Fang Wen}, {and}
  \bibinfo{person}{Baining Guo}.} \bibinfo{year}{2020}\natexlab{}.
\newblock \showarticletitle{Face X-ray for More General Face Forgery
  Detection}.
\newblock \bibinfo{journal}{\emph{The Conference on Computer Vision and Pattern
  Recognition (CVPR)}} (\bibinfo{year}{2020}).
\newblock


\bibitem[\protect\citeauthoryear{Li, Chang, Farid, and Lyu}{Li
  et~al\mbox{.}}{2018}]%
        {li2018ictu}
\bibfield{author}{\bibinfo{person}{Yuezun Li}, \bibinfo{person}{Ming-Ching
  Chang}, \bibinfo{person}{Hany Farid}, {and} \bibinfo{person}{Siwei Lyu}.}
  \bibinfo{year}{2018}\natexlab{}.
\newblock \showarticletitle{In ictu oculi: Exposing ai generated fake face
  videos by detecting eye blinking}.
\newblock \bibinfo{journal}{\emph{IEEE Workshop on Information Forensics and
  Security (WIFS)}} (\bibinfo{year}{2018}).
\newblock


\bibitem[\protect\citeauthoryear{Li and Lyu}{Li and Lyu}{2019}]%
        {li2018exposing}
\bibfield{author}{\bibinfo{person}{Yuezun Li} {and} \bibinfo{person}{Siwei
  Lyu}.} \bibinfo{year}{2019}\natexlab{}.
\newblock \showarticletitle{Exposing deepfake videos by detecting face warping
  artifacts}.
\newblock \bibinfo{journal}{\emph{Workshop on Media Forensics (in conjuction
  with CVPR)}} (\bibinfo{year}{2019}).
\newblock


\bibitem[\protect\citeauthoryear{Liu, Luo, Wang, and Tang}{Liu
  et~al\mbox{.}}{2015}]%
        {liu2015deep}
\bibfield{author}{\bibinfo{person}{Ziwei Liu}, \bibinfo{person}{Ping Luo},
  \bibinfo{person}{Xiaogang Wang}, {and} \bibinfo{person}{Xiaoou Tang}.}
  \bibinfo{year}{2015}\natexlab{}.
\newblock \showarticletitle{Deep learning face attributes in the wild}. In
  \bibinfo{booktitle}{\emph{International Conference on Computer Vision
  (ICCV)}}.
\newblock


\bibitem[\protect\citeauthoryear{Matern, Riess, and Stamminger}{Matern
  et~al\mbox{.}}{2019}]%
        {matern2019exploiting}
\bibfield{author}{\bibinfo{person}{Falko Matern}, \bibinfo{person}{Christian
  Riess}, {and} \bibinfo{person}{Marc Stamminger}.}
  \bibinfo{year}{2019}\natexlab{}.
\newblock \showarticletitle{Exploiting visual artifacts to expose deepfakes and
  face manipulations}. In \bibinfo{booktitle}{\emph{2019 IEEE Winter
  Applications of Computer Vision Workshops (WACVW)}}.
\newblock


\bibitem[\protect\citeauthoryear{Nguyen, Yamagishi, and Echizen}{Nguyen
  et~al\mbox{.}}{2019b}]%
        {nguyen2019capsule}
\bibfield{author}{\bibinfo{person}{Huy~H Nguyen}, \bibinfo{person}{Junichi
  Yamagishi}, {and} \bibinfo{person}{Isao Echizen}.}
  \bibinfo{year}{2019}\natexlab{b}.
\newblock \showarticletitle{Capsule-Forensics: Using Capsule Networks to Detect
  Forged Images and Videos}. In \bibinfo{booktitle}{\emph{International
  Conference on Acoustics, Speech, and Signal Processing (ICASSP)}}.
\newblock


\bibitem[\protect\citeauthoryear{Nguyen, Nguyen, Nguyen, Nguyen, and
  Nahavandi}{Nguyen et~al\mbox{.}}{2019a}]%
        {nguyen2019deep}
\bibfield{author}{\bibinfo{person}{Thanh~Thi Nguyen}, \bibinfo{person}{Cuong~M
  Nguyen}, \bibinfo{person}{Dung~Tien Nguyen}, \bibinfo{person}{Duc~Thanh
  Nguyen}, {and} \bibinfo{person}{Saeid Nahavandi}.}
  \bibinfo{year}{2019}\natexlab{a}.
\newblock \showarticletitle{Deep Learning for Deepfakes Creation and
  Detection}.
\newblock \bibinfo{journal}{\emph{arXiv preprint arXiv:1909.11573}}
  (\bibinfo{year}{2019}).
\newblock


\bibitem[\protect\citeauthoryear{Radford, Metz, and Chintala}{Radford
  et~al\mbox{.}}{2016}]%
        {radford2015unsupervised}
\bibfield{author}{\bibinfo{person}{Alec Radford}, \bibinfo{person}{Luke Metz},
  {and} \bibinfo{person}{Soumith Chintala}.} \bibinfo{year}{2016}\natexlab{}.
\newblock \showarticletitle{Unsupervised representation learning with deep
  convolutional generative adversarial networks}.
\newblock \bibinfo{journal}{\emph{The International Conference on Learning
  Representations (ICLR)}} (\bibinfo{year}{2016}).
\newblock


\bibitem[\protect\citeauthoryear{Rahmouni, Nozick, Yamagishi, and
  Echizen}{Rahmouni et~al\mbox{.}}{2017}]%
        {rahmouni2017distinguishing}
\bibfield{author}{\bibinfo{person}{Nicolas Rahmouni}, \bibinfo{person}{Vincent
  Nozick}, \bibinfo{person}{Junichi Yamagishi}, {and} \bibinfo{person}{Isao
  Echizen}.} \bibinfo{year}{2017}\natexlab{}.
\newblock \showarticletitle{Distinguishing computer graphics from natural
  images using convolution neural networks}. In \bibinfo{booktitle}{\emph{2017
  IEEE Workshop on Information Forensics and Security (WIFS)}}.
\newblock


\bibitem[\protect\citeauthoryear{Ronneberger, Fischer, and Brox}{Ronneberger
  et~al\mbox{.}}{2015}]%
        {ronneberger2015u}
\bibfield{author}{\bibinfo{person}{Olaf Ronneberger}, \bibinfo{person}{Philipp
  Fischer}, {and} \bibinfo{person}{Thomas Brox}.}
  \bibinfo{year}{2015}\natexlab{}.
\newblock \showarticletitle{U-net: Convolutional networks for biomedical image
  segmentation}. In \bibinfo{booktitle}{\emph{International Conference on
  Medical image computing and computer-assisted intervention}}. Springer.
\newblock


\bibitem[\protect\citeauthoryear{Rossler, Cozzolino, Verdoliva, Riess, Thies,
  and Nie{\ss}ner}{Rossler et~al\mbox{.}}{2019}]%
        {rossler2019faceforensics}
\bibfield{author}{\bibinfo{person}{Andreas Rossler}, \bibinfo{person}{Davide
  Cozzolino}, \bibinfo{person}{Luisa Verdoliva}, \bibinfo{person}{Christian
  Riess}, \bibinfo{person}{Justus Thies}, {and} \bibinfo{person}{Matthias
  Nie{\ss}ner}.} \bibinfo{year}{2019}\natexlab{}.
\newblock \showarticletitle{Faceforensics++: Learning to detect manipulated
  facial images}.
\newblock \bibinfo{journal}{\emph{International Conference on Computer Vision
  (ICCV)}} (\bibinfo{year}{2019}).
\newblock


\bibitem[\protect\citeauthoryear{Selvaraju, Cogswell, Das, Vedantam, Parikh,
  and Batra}{Selvaraju et~al\mbox{.}}{2017}]%
        {selvaraju2017grad}
\bibfield{author}{\bibinfo{person}{Ramprasaath~R Selvaraju},
  \bibinfo{person}{Michael Cogswell}, \bibinfo{person}{Abhishek Das},
  \bibinfo{person}{Ramakrishna Vedantam}, \bibinfo{person}{Devi Parikh}, {and}
  \bibinfo{person}{Dhruv Batra}.} \bibinfo{year}{2017}\natexlab{}.
\newblock \showarticletitle{Grad-cam: Visual explanations from deep networks
  via gradient-based localization}. In \bibinfo{booktitle}{\emph{International
  Conference on Computer Vision (ICCV)}}.
\newblock


\bibitem[\protect\citeauthoryear{Simonyan and Zisserman}{Simonyan and
  Zisserman}{2015}]%
        {simonyan2014very}
\bibfield{author}{\bibinfo{person}{Karen Simonyan} {and}
  \bibinfo{person}{Andrew Zisserman}.} \bibinfo{year}{2015}\natexlab{}.
\newblock \showarticletitle{Very deep convolutional networks for large-scale
  image recognition}.
\newblock \bibinfo{journal}{\emph{The International Conference on Learning
  Representations (ICLR)}} (\bibinfo{year}{2015}).
\newblock


\bibitem[\protect\citeauthoryear{Thies, Zollhofer, Stamminger, Theobalt, and
  Nie{\ss}ner}{Thies et~al\mbox{.}}{2016}]%
        {thies2016face2face}
\bibfield{author}{\bibinfo{person}{Justus Thies}, \bibinfo{person}{Michael
  Zollhofer}, \bibinfo{person}{Marc Stamminger}, \bibinfo{person}{Christian
  Theobalt}, {and} \bibinfo{person}{Matthias Nie{\ss}ner}.}
  \bibinfo{year}{2016}\natexlab{}.
\newblock \showarticletitle{Face2face: Real-time face capture and reenactment
  of rgb videos}. In \bibinfo{booktitle}{\emph{The Conference on Computer
  Vision and Pattern Recognition (CVPR)}}.
\newblock


\bibitem[\protect\citeauthoryear{Tolosana, Vera-Rodriguez, Fierrez, Morales,
  and Ortega-Garcia}{Tolosana et~al\mbox{.}}{2020}]%
        {tolosana2020deepfakes}
\bibfield{author}{\bibinfo{person}{Ruben Tolosana}, \bibinfo{person}{Ruben
  Vera-Rodriguez}, \bibinfo{person}{Julian Fierrez}, \bibinfo{person}{Aythami
  Morales}, {and} \bibinfo{person}{Javier Ortega-Garcia}.}
  \bibinfo{year}{2020}\natexlab{}.
\newblock \showarticletitle{DeepFakes and Beyond: A Survey of Face Manipulation
  and Fake Detection}.
\newblock \bibinfo{journal}{\emph{arXiv preprint arXiv:2001.00179}}
  (\bibinfo{year}{2020}).
\newblock


\bibitem[\protect\citeauthoryear{Wu, AbdAlmageed, and Natarajan}{Wu
  et~al\mbox{.}}{2019}]%
        {wu2019mantra}
\bibfield{author}{\bibinfo{person}{Yue Wu}, \bibinfo{person}{Wael AbdAlmageed},
  {and} \bibinfo{person}{Premkumar Natarajan}.}
  \bibinfo{year}{2019}\natexlab{}.
\newblock \showarticletitle{ManTra-Net: Manipulation Tracing Network for
  Detection and Localization of Image Forgeries With Anomalous Features}. In
  \bibinfo{booktitle}{\emph{The Conference on Computer Vision and Pattern
  Recognition (CVPR)}}.
\newblock


\bibitem[\protect\citeauthoryear{Xuan, Peng, Dong, and Wang}{Xuan
  et~al\mbox{.}}{2019}]%
        {xuan2019generalization}
\bibfield{author}{\bibinfo{person}{Xinsheng Xuan}, \bibinfo{person}{Bo Peng},
  \bibinfo{person}{Jing Dong}, {and} \bibinfo{person}{Wei Wang}.}
  \bibinfo{year}{2019}\natexlab{}.
\newblock \showarticletitle{On the generalization of GAN image forensics}.
\newblock \bibinfo{journal}{\emph{arXiv preprint arXiv:1902.11153}}
  (\bibinfo{year}{2019}).
\newblock


\bibitem[\protect\citeauthoryear{Yu, Lin, Yang, Shen, Lu, and Huang}{Yu
  et~al\mbox{.}}{2018}]%
        {yu2018generative}
\bibfield{author}{\bibinfo{person}{Jiahui Yu}, \bibinfo{person}{Zhe Lin},
  \bibinfo{person}{Jimei Yang}, \bibinfo{person}{Xiaohui Shen},
  \bibinfo{person}{Xin Lu}, {and} \bibinfo{person}{Thomas~S Huang}.}
  \bibinfo{year}{2018}\natexlab{}.
\newblock \showarticletitle{Generative image inpainting with contextual
  attention}. In \bibinfo{booktitle}{\emph{The Conference on Computer Vision
  and Pattern Recognition (CVPR)}}.
\newblock


\bibitem[\protect\citeauthoryear{Zhou, Khosla, Lapedriza, Oliva, and
  Torralba}{Zhou et~al\mbox{.}}{2016}]%
        {zhou2016learning}
\bibfield{author}{\bibinfo{person}{Bolei Zhou}, \bibinfo{person}{Aditya
  Khosla}, \bibinfo{person}{Agata Lapedriza}, \bibinfo{person}{Aude Oliva},
  {and} \bibinfo{person}{Antonio Torralba}.} \bibinfo{year}{2016}\natexlab{}.
\newblock \showarticletitle{Learning deep features for discriminative
  localization}. In \bibinfo{booktitle}{\emph{The Conference on Computer Vision
  and Pattern Recognition (CVPR)}}.
\newblock


\end{thebibliography}

\end{document}